\documentclass[letterpaper]{article} 
\usepackage[]{aaai2026}  
\usepackage{times}  
\usepackage{helvet}  
\usepackage{courier}  
\usepackage[hyphens]{url}  
\usepackage{graphicx} 
\usepackage{natbib}  
\usepackage{caption} 
\usepackage{algorithm}
\usepackage{algorithmic}
\usepackage{amsmath,amsfonts}
\usepackage{booktabs}
\usepackage{multirow}
\usepackage{xcolor}
\usepackage{newfloat}
\usepackage{listings} 

\nocopyright

\DeclareCaptionStyle{ruled}{labelfont=normalfont,labelsep=colon,strut=off} 
\floatstyle{ruled}
\newfloat{listing}{tb}{lst}{}
\floatname{listing}{Listing}
\title{Cross3DReg: Towards a Large-scale Real-world Cross-source Point Cloud Registration Benchmark}
\author{
    Zongyi Xu,
    Zhongpeng Lang,
    Yilong Chen,
    Shanshan Zhao,
    Xiaoshui Huang, Yifan Zuo,
    Yan Zhang,
    Qianni Zhang,
    Xinbo Gao
}
\affiliations{

}

\usepackage{bibentry}

\begin{document}

\maketitle

\begin{abstract}

Cross-source point cloud registration, which aims to align point cloud data from different sensors, is a fundamental task in 3D vision. However, compared to the same-source point cloud registration, cross-source registration faces two core challenges: the lack of publicly available large-scale real-world datasets for training the deep registration models, and the inherent differences in point clouds captured by multiple sensors. The diverse patterns induced by the sensors pose great challenges in robust and accurate point cloud feature extraction and matching, which negatively influence the registration accuracy. To advance research in this field, we construct Cross3DReg, the currently largest and real-world multi-modal cross-source point cloud registration dataset, which is collected by a rotating mechanical lidar and a hybrid semi-solid-state lidar, respectively. Moreover, we design an overlap-based cross-source registration framework, which utilizes unaligned images to predict the overlapping region between source and target point clouds, effectively filtering out redundant points in the irrelevant regions and significantly mitigating the interference caused by noise in non-overlapping areas.
Then, a visual-geometric attention guided matching module is proposed to enhance the consistency of cross-source point cloud features by fusing image and geometric information to establish reliable correspondences and ultimately achieve accurate and robust registration.
Extensive experiments show that our method achieves state-of-the-art registration performance. Our framework reduces the relative rotation error (RRE) and relative translation error (RTE) by $63.2\%$ and $40.2\%$, respectively, and improves the registration recall (RR) by $5.4\%$, which validates its effectiveness in achieving accurate cross-source registration. The code and dataset will be released at \color{magenta} {https://github.com/bestmyselfllll/Cross3DReg}.

\end{abstract}


\section{Introduction}
Cross-source point cloud registration \cite{zhao2025cross,huang2021comprehensive} is a fundamental task in 3D vision, which plays an important role in robot navigation, remote sensing mapping, and localization for autonomous driving \cite{xiong2024speal}. The goal of cross-source point cloud registration is to align point clouds acquired from different sensors to construct complete 3D scenes \cite{huang2023} or estimate the robot location on maps \cite{wang2025end}.

\begin{figure}[!]
    \centering
    \includegraphics[width=\linewidth]{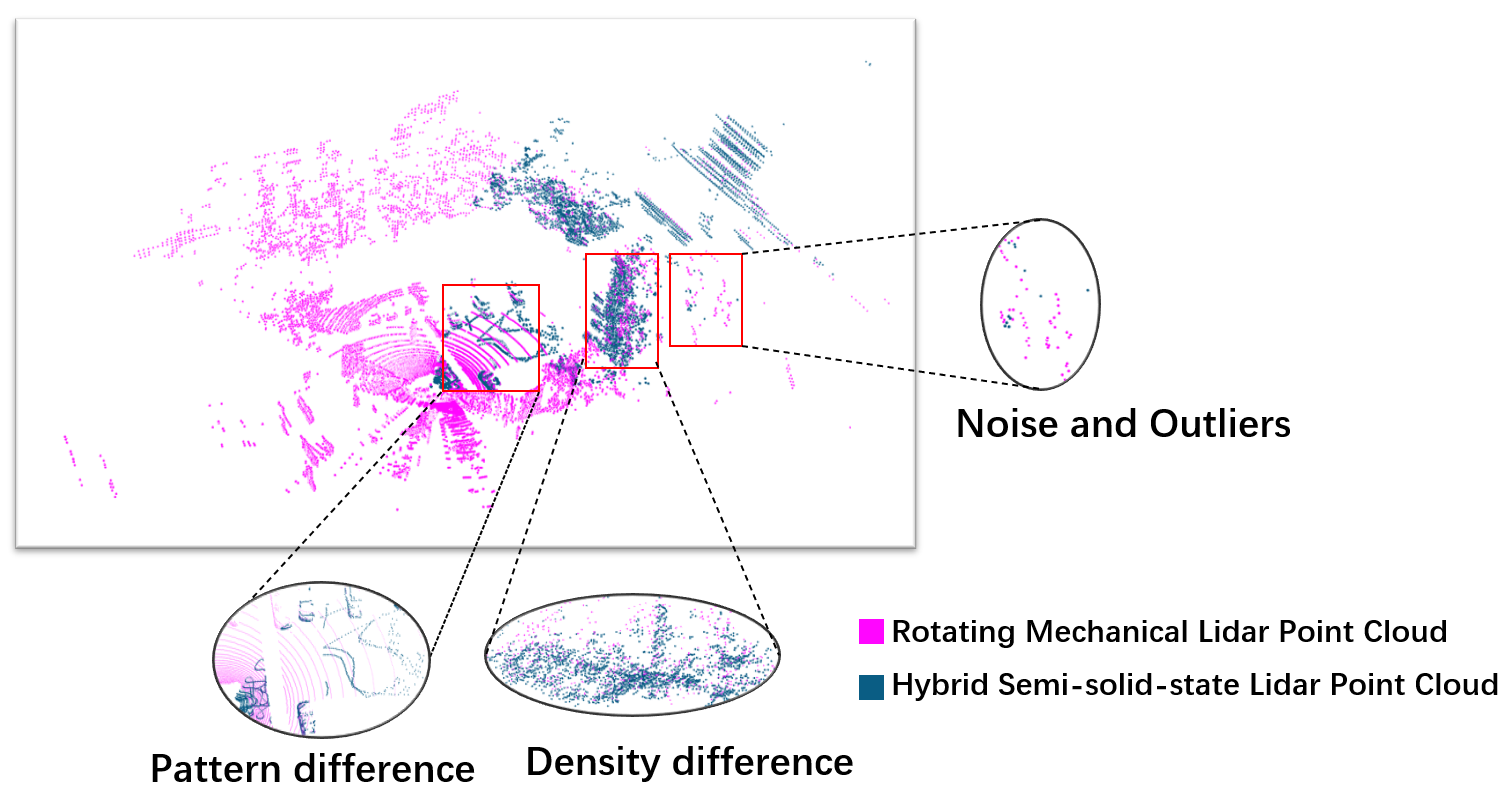}
    \caption{Challenges like differences in structural pattern and density, and realistic noise and outliers presented in real-world cross-source point clouds. 
    }
    \label{fig:problems}
\end{figure}
Compared to the same-source point cloud registration, advancements in cross-source registration are relatively slow for two main reasons. Firstly, there is a severe lack of public benchmark datasets that possess sufficient cross-source point cloud pairs. 
Existing public datasets, such as 3DCGS \cite{huang2021comprehensive} and KITTI-CrossSource \cite{xiong2024speal}, exhibit notable limitations. 3DCGS \cite{huang2021comprehensive} provides point clouds of Kinect-SFM indoor scenes. Its data scale is relatively small and insufficient for training deep registration models. The KITTI CrossSource dataset \cite{xiong2024speal} consists of lidar scans and reconstructed point clouds from sequences using MonoRec \cite{wimbauer2021monorec}. In the two cross-source point cloud datasets, the source or target point clouds are mainly synthesised with image sequences, not captured with real sensors. Secondly, as shown in Figure \ref{fig:problems}, point clouds scanned from different types of real scanners exhibit significant variance in data density and structural pattern. For example, point clouds captured by rotating LiDAR typically display sparse ring-like structures, whereas point clouds scanned by semi-solid-state LiDAR are often fan-shaped. In addition, point clouds acquired from different sensors vary considerably in terms of noise levels, outlier distributions, and missing regions. These discrepancies can easily lead to a large number of mismatched features, which in turn severely degrade registration performance. Moreover, challenges from the real world, like inherent noise, outliers from capturing sensors and different data structure patterns, are not presented in the currently available cross-source datasets.   
Facing these real-world cross-source point clouds, the accuracy of most existing same-source point cloud registration methods \cite{yu2023rotation,ren2024multi,mu2024colorpcr,jiang2025zero} often deteriorates significantly due to different levels of noise and variance in density and structural patterns. Recent cross-source point cloud registration methods \cite{zhao2025cross, xiong2024speal,huang2023cross,huang2017systematic} are also proposed to improve registration accuracy. However, they are only designed for synthetic cross-source settings like Kinect-SFM, not real-world cross-source datasets.


To address the above challenges, we firstly construct a large-scale real-world cross-source point cloud registration dataset named Cross3DReg. The source point clouds are captured by a hybrid semi-solid-state Lidar, and the target point clouds are acquired with a 64-line rotating mechanical Lidar. We also capture the front views of the scenes with an RGB camera when the point clouds are obtained. RGB images only show the views and are not aligned with the point clouds.
To mitigate interference from outliers and noise in point cloud matching, as well as the feature inconsistency arising from differences in cross-source point cloud density and structural pattern,
we also propose an overlap-based cross-source point cloud registration method. Inspired by ImLoveNet \cite{chen2022imlovenet}, we first predict the overlap regions between the source and target point clouds. Then, a dual-modal encoder is designed that effectively fuses the global features of RGB images with the coarse-grained geometric features of point clouds. This allows for the prediction of the overlap region masks between point clouds and images.
By regarding the overlap region as the intersection of the source and target point clouds within the image's field of view, we can effectively filter out redundant points and noise from non-overlapping areas.
We also introduce a visual-geometric attention-guided matching module. This module enhances the consistency of cross-source point cloud features by fusing visual and geometric information adaptively with an attention mechanism, thereby establishing reliable correspondences.

Our main contributions are listed below:
\begin{itemize}
\item To the best of our knowledge, Cross3DReg is the first large-scale real-world cross-source point cloud registration dataset. It includes $13,231$ point cloud pairs where different levels of noise, outliers, densities, and structural patterns are presented. Images showing common views between source and target point clouds are also collected. The dataset and code will be released. 
\item An overlap-based cross-source point cloud registration method is proposed to achieve accurate registration by predicting the overlap region with the help of unaligned images and ignoring redundant points and noise. 
\item To achieve accurate feature matching within the overlap region, an attention mechanism is utilised to adaptively fuse geometric features of point clouds with unaligned image features, enhancing feature consistency between cross-source points.
    
\end{itemize}

\section{Related Work}
\textbf{Same-source point could registration}. Currently, same-source point cloud registration methods can be categorized into three classes. The first class is traditional iterative optimization-based approaches \cite{besl1992method,rusinkiewicz2001efficient,segal2009generalized}. However, these methods often suffer from a significant drop in registration accuracy when dealing with noisy or structurally complex scenes. The second class is correspondence-based point cloud registration methods \cite{choy2019fully,wang2022you,yu2023rotation}. Early approaches primarily rely on handcrafted feature descriptors \cite{rusu2008aligning,rusu2009fast} to establish point-wise correspondences. With the rapid development of deep learning in point cloud registration, a series of deep learning-based feature extraction models \cite{bai2020d3feat,ao2021spinnet} have been proposed, enabling more accurate and robust registration. Nevertheless, these methods still experience notable performance degradation when applied to regions with low overlap or a large number of outliers. To address these challenges, coarse-to-fine registration strategies \cite{yu2021cofinet, qin2023geotransformer} have recently been adopted, showing accurate registration results under conditions of low overlap and high noise levels. The third class is end-to-end point cloud registration methods \cite{xu2021omnet,zhang2022end,lu2019deepvcp}. Unlike conventional two-stage registration frameworks, end-to-end approaches directly utilise deep neural networks to predict rigid transformations between point clouds without explicitly establishing point correspondences, thereby improving overall registration efficiency. Moreover, since raw point cloud data contains only geometric information, recent multimodal point cloud registration methods \cite{zhang2022pcr,xu2024igreg,xu2025s2reg} attempt to enhance feature discriminability by incorporating additional modalities (like colour, semantics, texture), allowing for more reliable correspondences. However, such methods could potentially fail in the cross-source point cloud registration task.

\textbf{Cross-Source point cloud Registration}. The core challenge in cross-source point cloud registration lies in addressing the significant discrepancies introduced by different types of sensors. Compared to same-source registration, cross-source point clouds exhibit a gap in density and pattern distribution, and are more susceptible to outliers. Traditional methods \cite{huang2017coarse,huang2019fast} are not designed to cope with these problems. To address the specific challenges posed by cross-source data, in recent years, mainstream cross-source point cloud registration methods \cite{ma2024ff,xiong2024speal,zhao2025cross} widely adopt a coarse-to-fine strategy. Correspondences are established by learning consistent deep features between the cross-source point clouds, thereby achieving robust registration. However, the advancement of this field is constrained by limited datasets; the efficacy of current approaches \cite{zhao2024vrhcf,zhao2025cross} has primarily been validated on cross-source datasets where the source/target point clouds are synthesised with images, not real-world cross-source datasets. Therefore, developing cross-source point cloud registration data using real-world sensors represents a key future research objective in this area.

\section{The Cross3DReg Dataset}

To facilitate the development of cross-source point cloud registration, we introduce a large-scale real-world Cross3DReg dataset that contains $13,231$ point cloud pairs captured by a rotating mechanical LiDAR and a hybrid semi-solid-state LiDAR. Images are also collected when the calibration between the RGB camera and the Lidars is unavailable. 
The data acquisition platform is a custom-built Unmanned Ground Vehicle (UGV), equipped with a 64-beam spinning mechanical LiDAR, a hybrid solid-state LiDAR, and an RGB camera. Specifically, the source point clouds are captured by the hybrid semi-solid-state LiDAR, while the target point clouds are acquired by the rotating mechanical LiDAR. The RGB camera captures front-view images of the scene to provide auxiliary visual information.  More details about the Cross3DReg dataset, please refer to the \textit{supplementary material}. 

\begin{table*}[h]
    \centering
    \small
    \begin{tabular}{cccccccccc}
\toprule
\textbf{Dataset} & \textbf{Scenes} & \textbf{Sensors} & \textbf{Pairs} & \textbf{Img.} & \textbf{Po.} & \textbf{Rno.} & \textbf{Rdp.} & \textbf{Dens.}  &\textbf{Open.}\\
\midrule
3DCGS & Indoors & Kinect, Lidar, SFM & 202 & $\times$ & $\checkmark$ & $\times$ & $\times$ & $\checkmark$  &$\checkmark$  \\
KITTI-CrossSource & outdoors & Lidar, SFM & $-^{*}$ & $\checkmark$ & $\checkmark$ & $\times$ & $\times$ & $\checkmark$  &$\times$ \\
Cross3DReg & outdoors & RL,HL & 13231 & $\checkmark$ & $\checkmark$ & $\checkmark$ & $\checkmark$ & $\checkmark$  &$\checkmark$  \\
\bottomrule
\end{tabular}
    \caption{The comparison of cross-source point cloud datasets. Img: RGB images. Po: Partial overlap. Rno: Real-world noise and outliers. Rdp: Real difference of structural pattern. Dens: Density difference. Open: Open source. RL: Rotating mechanical Lidar. HL: Hybrid semi-solid-state Lidar. $-^{*}$: As the KITTI-CrossSource is unreleased, the number of point cloud pairs is unclear.}
    \label{tab:datasets}
\end{table*}

\section{Method}
\begin{figure*}[ht]
\centering
\includegraphics[width=1\textwidth]{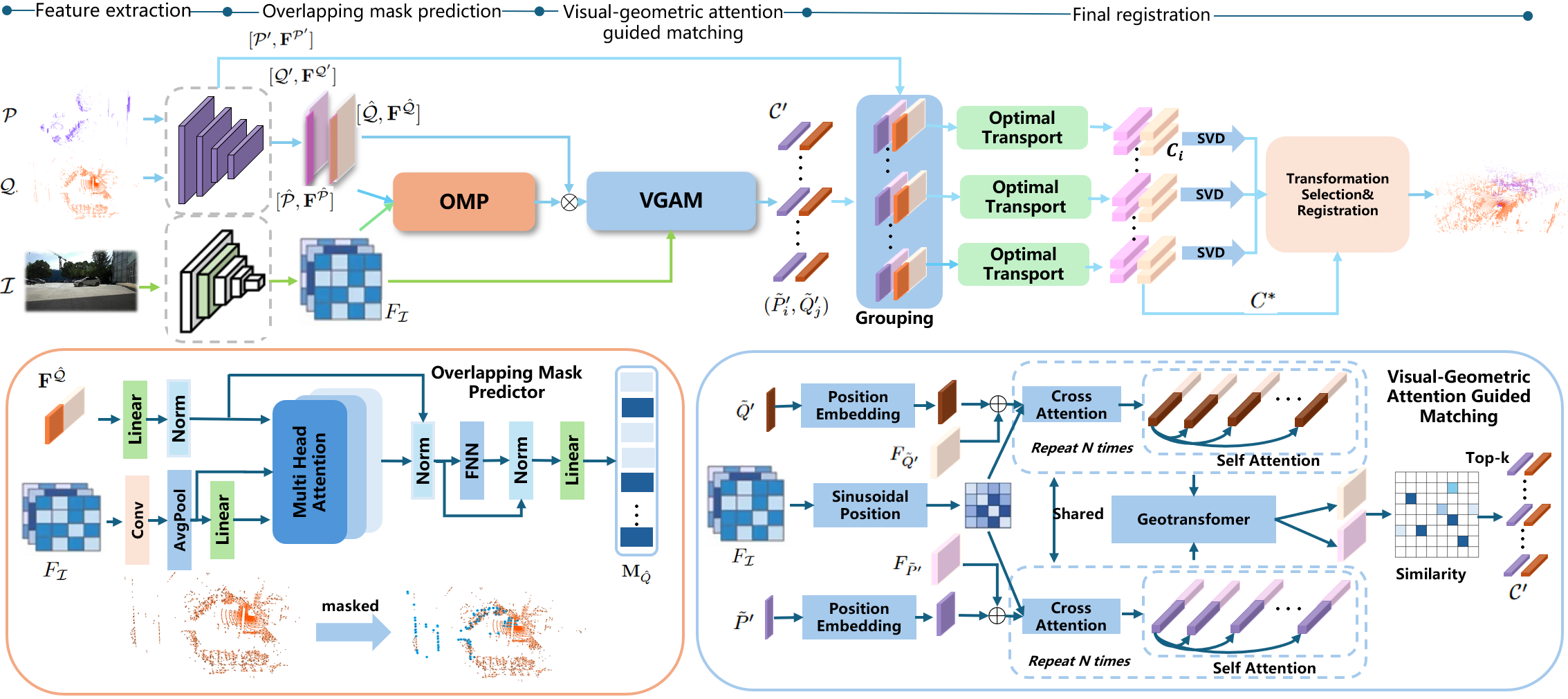} 
\caption{Our Cross3DReg consists of four parts.
Firstly, a pair of cross-source point clouds $\mathcal{P}$ and $\mathcal{Q}$, and an unaligned image $\mathcal{I}$ partially overlapping with them, are input to corresponding feature encoders to obtain the downsampled superpoints and feature descriptors, respectively. Next, we fuse the extracted image features with the features of the two point clouds, respectively, and predict the respective overlapping region masks between the image and point clouds accordingly. Afterwards,
we input the superpoints in the overlap region and their features into the vision-geometric guided matching module to establish superpoint correspondences $\mathcal{C}'$. Finally, we propagate these correspondences between the superpoints into the original dense point cloud to establish the final correspondences $\mathcal{C}^{*}$, and input them into the specified pose estimator to obtain the transformation matrix.}

\label{fig:frame}
\end{figure*}

\textbf{Problem Statement}. The cross-source point cloud registration is formulated as follows. Given the source point cloud $\mathcal{P} = \{p_i \in \mathbb{R}^3 | i=1, \dots, N\}$ and target point cloud $\mathcal{Q} = \{q_j \in \mathbb{R}^3 | j=1, \dots, M\}$, we estimate an optimal rigid transformation matrix $\mathcal{T} = \{\mathbf{R,t}$\} to align $\mathcal{P}$ and $\mathcal{Q}$, where $\mathbf{R} \in {SO}(3) $ is the rotation matrix and $t \in \mathbb{R}^3$ is the translation vector. The optimisation goal can be formulated as:
\begin{equation}
    \mathcal{T} = \arg \min_{R,t} \sum_{(p_i,q_j) \in C^*} \|\mathbf{R}\mathbf{p}_i + \mathbf{t} - \mathbf{q}_j\|_2,
\end{equation}
where $\mathcal{C}^*$ represents the correspondences between $\mathcal{P}$ and $\mathcal{Q}$ and the $\|.\|_2$ denotes the Euclidean paradigm.

For accurate cross-source registration, we propose an overlap-based cross-source point registration method to establish the correspondence between point clouds. 
As shown in Figure \ref{fig:frame}, the framework contains four phases: 1) Feature extraction. The features of Image $\mathcal{I}$ and point clouds $\mathcal{P}$ and $\mathcal{Q}$ are extracted through a two-branch network. The point cloud branch downsamples the original point clouds to acquire superpoints  $\mathcal{\hat{P}}$ and $\mathcal{\hat{Q}}$. The corresponding features are also extracted; 2) Overlapping mask prediction (OMP). Although images are not aligned with point clouds, they contain the common views of the source and target point clouds. Thus, the coarse-grained superpoint features are firstly fused with image features to predict overlap region masks between images and point clouds. Both the overlap masks of the source and target point clouds are located. The following matching step is only performed on the overlap region. 3) Visual-geometric attention guided superpoint matching (VGAM). In the overlap region, visually enhanced superpoint features are utilized for similarity computation, and robust coarse-grained correspondences are established. 4) Point matching and registration. Based on the superpoint matching results, precise point-level correspondences are obtained by local point cloud feature matching. Based on the point correspondences, 
transformations are estimated using the pose estimators.

\subsection{Feature Extraction}

Due to the significant density variations and a large number of points, we first preprocess the raw point clouds using a voxel-based downsampling method (voxel size = 0.25) before feeding them into the feature extractor. 
The downsampled source point cloud $\mathcal{P}$ and the target point cloud $\mathcal{Q}$ are then put into the KPConv-FPN backbone network \cite{thomas2019kpconv}, which can extract the point cloud features at different scales. At the coarsest scale, we obtain the superpoint sets $\mathcal{\hat{P}}$ and $\mathcal{\hat{Q}}$. Their corresponding features are $\mathbf{F}^{\mathcal{\hat{P}}}\in \mathbb{R}^{|\hat{P}| \times \hat{d}} $ and $\mathbf{F}^{\mathcal{\hat{Q}}}\in \mathbb{R}^{|\hat{Q}|\times \hat{d}}$,
$\hat{d}$ is the feature dimension of superpoints at the coarsest level. Points at the densest level are denoted as $\mathcal{P}', \mathcal{Q}'$ and its corresponding features are $\mathbf{F}^{\mathcal{P}'} \in \mathbb{R}^{|\mathcal{P}'| \times d'}$ and $\mathbf{F}^{\mathcal{Q}'} \in \mathbb{R}^{|\mathcal{Q}'| \times d'}$, $d'$ is the corresponding feature dimension. 

For image feature extraction, we employ a U-Net backbone network with residual connections to process the intermediate unaligned images. 
Given an input image $\mathcal{I} \in \mathbb{R}^{H \times W} $,  its feature is represented as ${F}_{\mathcal{I}} \in \mathbb{R}^{H \times W \times d} $.

\subsection{Overlapping Mask Predictor Module}

As various point densities, prevalent noise, outliers and different structural patterns exist in the cross-source point cloud registration, traditional point features-based overlapping region estimation is inaccurate. Since the captured images in our dataset contain the common views, even though the images and point clouds are not aligned, we can utilise images to acquire rough overlapping regions between the source and target point clouds.

Thus, we propose an overlapping mask prediction module using unaligned images. We first extract overlap regions between images and source point clouds, and those between images and target point clouds. As the images show, we consider the point cloud-image overlap region a good reference for the overlap between point clouds. 
Since the calibration information is unavailable, images and point clouds are unaligned. We first perform image and point cloud feature alignment.
The image features $\mathbf{F}_{\mathcal{I}} \in \mathbb{R}^{H \times W \times d} $ and superpoint features $\mathbf{F}^{\mathcal{\hat{Q}}}\in \mathbb{R}^{|\hat{Q}|\times \hat{d}}$ are mapped to a unified dimensional space by linear projection to obtain aligned features. The aligned image features and the superpoint features of point clouds are represented as $\widehat{\textbf{F}}^{\mathcal{\hat{Q}}}$ and $\widehat{\textbf{F}}_{\mathcal{I}}$.
Then, we use the multi-attention mechanism \cite{vaswani2017attention} to carry out the cross-modal feature fusion. Finally, the fused features are processed by a residual feed-forward network (FFN) with layer normalization.  The overlap probability of superpoints is output via an MLP net. Taking the prediction of the overlap mask for the target point cloud $\mathcal{\mathbf{Q}}$ as an example, the process can be formalized as follows.

\begin{equation}
    \mathbf{F}_{fuse} = MultiHeadAttn(\widehat{\mathbf{F}}^{\mathcal{\hat{Q}}},\widehat{\mathbf{F}}_{\mathcal{I}}).
\end{equation}

Given the fused image and point cloud features, the probability of each point belonging to an overlapping mask, denotes as $\mathbf{P}_{overlap}^{\hat{Q}}$, can be estimated as:

\begin{equation}
    \mathbf{P}_{overlap}^{\hat{Q}}= \sigma(MLP(\mathbf{F}_{fuse} + ( FFN(\mathbf{F}_{fuse} + \widehat{F}^{\mathcal{\hat{Q}}})))),
\end{equation}


\begin{equation}
    \mathbf{M}_{\hat{Q}_i} = \begin{cases}
        1 &   \text{if } {\mathbf{P}_{overlap}^{\hat{Q}_i} > \lambda}\\
        0 &  \text{otherwise }
        
    \end{cases},
\end{equation}
where $\mathbf{M}_{\hat{Q}} \in \{0,1\}$ is a binary mask vector, $\sigma$ represents the sigmoid function, and the $\lambda$ denotes the confidence threshold, default is $0.5$.

\subsection{
Visual-Geometric Attention Guided Matching
}

To effectively incorporate visual context into geometric features for point cloud registration, we introduce the visual-geometric attention guided
matching module
to leverage both visual and geometric information to enhance the consistency of superpoint features between point cloud of different sources. The core of our approach is a two-stage attention mechanism. First, the visual cross-attention mechanism fuses visual context from image features into the superpoint features. Subsequently, a geometric self-attention mechanism\cite{qin2023geotransformer} refines these fused features to capture global geometric relationships within the point cloud.

Let's take the target point cloud $\mathcal{Q}$ as an example. We first use a predicted overlap mask $\mathbf{M}_{\hat{Q}}$ to select a subset of superpoints $\mathbf{\tilde{Q}}'$ and their corresponding features $\mathbf{F}_{\tilde{Q}'}$ located in the overlapping region. The corresponding image features $\mathbf{F}_I$ are flattened into a vector. To provide the attention mechanism with spatial awareness, we introduce positional encoding, $\mathbf{F}_{pos}^{\tilde{Q}'}$ for superpoints and $\mathbf{F}_{pos}^{I}$ for image pixels.

The superpoint features, image features, and their respective positional encodings are projected into Query ($\mathbf{Q}_c$), Key ($\mathbf{K}_c$), and Value ($\mathbf{V}_c$) spaces using learnable linear matrices. The positional encodings are also projected to generate point cloud positional embeddings ($\mathbf{E}_c$) and image positional embeddings ($\mathbf{G}_c$). The process is formulated as:

\begin{align}
    \mathbf{Q}_c &= \mathbf{F}_{\tilde{Q}'} \mathbf{W}_{Q_c}, \quad \mathbf{K}_c = \mathbf{F}_I \mathbf{W}_{K_c}, \quad \mathbf{V}_c = \mathbf{F}_I \mathbf{W}_{V_c}, \\
    \mathbf{E}_c &= \mathbf{F}_{pos}^{\tilde{Q}'} \mathbf{W}_{E_c}, \quad \mathbf{G}_c = \mathbf{F}_{pos}^{I} \mathbf{W}_{G_c},
\end{align}
where $\mathbf{W}_{Q_c}, \mathbf{W}_{K_c}, \mathbf{W}_{V_c}$ are learnable projection matrices for the Query, Key, and Value, and $\mathbf{W}_{E_c}, \mathbf{W}_{G_c}$ are the projection matrices for their respective positional embeddings.

By integrating content and position information, we compute the cross-attention scores. These scores weigh the aggregation of the Value and image positional embeddings. The superpoint features are then updated via a residual connection:
\begin{align}
    \mathbf{Scores}_c &= \textit{softmax}\left(\frac{(\mathbf{Q}_c + \mathbf{E}_c)(\mathbf{K}_c + \mathbf{G}_c)^T}{\sqrt{d'}}\right) , \\
    \mathbf{F}'_{\tilde{Q}'} &= \text{Scores}_c (\mathbf{V}_c + \mathbf{G}_c) + \mathbf{F}_{\tilde{Q}'} ,
\end{align}
where $\mathbf{F}'_{\tilde{Q}'}$ denotes the updated superpoint features enriched with visual context information.

To enhance the global structural integrity of the features and mitigate potential noise, we employ a self-attention mechanism. This step promotes information propagation across the entire point cloud. The visually-enhanced features $\mathbf{F}'_{\tilde{Q}'}$ are linearly projected into a new set of Query ($\mathbf{Q}_s$), Key ($\mathbf{K}_s$), and Value ($\mathbf{V}_s$). Self-attention weights are then computed to update the features in a residual manner:
\begin{align}
    \mathbf{Scores}_s &= \textit{softmax}\left(\frac{\mathbf{Q}_s \mathbf{K}_s^T}{\sqrt{d'}}\right) , \\
    \mathbf{F}''_{\tilde{Q}'} &= \mathbf{Scores}_s \mathbf{V}_s + \mathbf{F}'_{\tilde{Q}'}.
\end{align}

Finally, combined with the geometric self-attention mechanism, we can maximize the descriptive power of the final features. After the feature enhancement process, we can obtain highly discriminative superpoint features, $\bar{F}_{\tilde{P}'}$ and $\mathbf{\bar{F}}_{\tilde{Q}'}$, for the source and target point clouds, respectively. We then construct a feature similarity matrix $\mathbf{Z}'$ by the following formulation:
\begin{equation}
    \mathbf{Z}'_{ij} = \exp\left(-\|\mathbf{\bar{F}}_{\tilde{P}'_i} - \mathbf{\bar{F}}_{\tilde{Q}'_j}\|_2^2\right),
\end{equation}
where $\mathbf{Z}'_{ij}$ measures the similarity between the $i$-th source superpoint $\mathbf{\tilde{P}}'_i$ and the $j$-th target superpoint $\mathbf{\tilde{Q}}'_j$. Finally, we apply dual normalization \cite{rocco2018neighbourhood, sun2021loftr} to the similarity matrix $\mathbf{Z}'$ and select the top-K entries with the highest scores to form the final set of superpoint correspondences $\mathcal{C}'=\{(\mathbf{\tilde{P}}'_i,\mathbf{\tilde{Q}}'_j)|\mathbf{\tilde{P}}'_i \in \mathcal{\hat{P}}, \mathbf{\tilde{Q}}'_j \in \mathcal{\hat{Q}}\}$.

\subsection{Point Matching and Registration}
After obtaining the superpoint correspondences in the overlapping regions, we employ a point-to-node grouping strategy \cite{yu2021cofinet} to further establish correspondences between the dense points. The core idea of this strategy is to assign dense points to their nearest neighboring superpoints based on spatial distance. Specifically, for a matched superpoint pair $(\mathbf{\tilde{P}}_i', \mathbf{\tilde{Q}}_j')$, we denote their corresponding dense point groups as $\mathbf{G}_{\tilde{P}_i'}$ and $\mathbf{G}_{\tilde{Q}_j'}$, and their feature groups as $\mathbf{G}_{i}^{\mathbf{F}^{\mathcal{P}'}}$ and $\mathbf{G}_{j}^{\mathbf{F}^{\mathcal{Q}'}}$, respectively. Based on the superpoint correspondence $(\mathbf{\tilde{P}}_i', \mathbf{\tilde{Q}}_j')$, we compute the similarity matrix between the feature groups $\mathbf{G}_{i}^{\mathbf{F}^{\mathcal{P}'}}$ and $\mathbf{G}_{j}^{\mathbf{F}^{\mathcal{Q}'}}$ as $\mathbf{S}_{l'} = \frac{\mathbf{G}_{i}^{\mathcal{P}'} (\mathbf{G}_{j}^{\mathbf{F}^{\mathcal{Q}'}})^T}{\tilde{d}}$, where $\tilde{d}$ denotes the feature dimension. To enhance matching robustness, we adopt the method from \cite{sarlin2020superglue} by adding slack terms, controlled by a learnable parameter $\alpha$, to the last row and column of the similarity matrix $\hat{\mathbf{S}}'$. Subsequently, we apply the Sinkhorn algorithm to find the optimal matching. After removing the slack terms, we select the top $K'$ matching pairs with the highest confidence scores to establish the group-level dense point correspondences $C_{l'}$. Finally, by aggregating the correspondences from all groups, we obtain the global correspondences $\mathcal{C}^{*} = \bigcup_{l=1}^{|C'|} \mathcal{C}_{l'}$. Based on the correspondences, we employ the LGR estimator \cite{qin2023geotransformer} to accurately estimate the transformation matrix.

\subsection{Loss Function}
Our loss function is composed of three components and can be expressed as: $\mathcal{L}_{total} = \mathcal{L}_{coarse} + \mathcal{L}_{fine} + \mathcal{L}_{mask}$.
Here, following the framework of GeoTrans \cite{qin2023geotransformer},
$\mathcal{L}_{coarse}$ and $\mathcal{L}_{fine}$ are the losses supervising the coarse-grained (superpoint) and fine-grained (point-level) matching, respectively.
For $L_{mask}$, we employ the Focal Loss \cite{lin2017focal}. Due to the lack of camera intrinsic and extrinsic parameters, which prevents the establishment of a projection relationship from the 3D point cloud to the 2D image, we adopt a mask generation strategy based on the point cloud overlap. Specifically, for a given source point cloud $\mathcal{P}$ and a target point cloud $\mathcal{Q}$, the ground-truth overlapping mask at the superpoint level, $\mathbf{M}_g^{i}$, is defined as:
\begin{equation}
    \mathbf{M}_{g}^{i}=\begin{cases} 
    1 & \text{if } \mathcal{\hat{P}}_i \text{ correspondent to } \mathcal{\hat{P}}_j \\ 
    0 & \text{otherwise} 
\end{cases},
\end{equation}

\begin{equation}
    \mathbf{p}_t^{i} = \begin{cases}
    p_i & \text{if}\quad\mathbf{M}_{g}^{i} =1 \\
    1-p_i & \text{otherwise}
    \end{cases},
\end{equation}

\begin{equation}
    \mathcal{L}_{mask} = \frac{1}{|\mathbf{M}_g|}\sum_{i}^{|\mathbf{M}_g|}-\alpha(1-\mathbf{p}_t^{i})^{\gamma}\log(\mathbf{p}_t^{i}),
\end{equation}
where $\mathbf{p}_i$ denotes the mask probability of model output. $\mathbf{p}_t^{i}$ denotes the probability that the mask value is true. $\gamma=2.0$ and $\alpha=0.25$ are the focusing parameter and balancing parameter, respectively.

\section{Experiments}

\textbf{Metrics}. We use the following metrics to evaluate methods: \textit{Relative Rotation Error} (\textbf{RRE}), \textit{Relative Translation Error} (\textbf{RTE}), \textit{Registration Recall} (\textbf{RR}), and \textit{Inlier Ratio} (\textbf{IR}). For the Cross3DReg dataset, the threshold for RR is defined as RRE $<$ $2^\circ$ and RTE $<$ $0.5$ m. The IR evaluates the quality of the point matching by calculating the proportion of corresponding points whose distances under the true transformation are below the threshold of $1.0$m.

\textbf{Implementation Details}. All experiments are implemented based on the PyTorch framework and trained on the NVIDIA RTX A6000 GPU with the following key parameters: initial learning rate is $10^{-4}$; Batch size is $1$, and the weight decay is $10^{-6}$. Our model is trained with Adam optimizer for $20$ epochs.

\begin{table}[!htbp]
    \centering 
    \begin{tabular}{@{}lccc@{}} 
        \toprule
        Method & RRE ($^\circ$) $\downarrow$ & RTE (m) $\downarrow$ & RR (\%) $\uparrow$\\
        \midrule
        ICP & 94.69 & 9.3 & 0.0\\
        FCGF & 94.70 & 9.0 & 0.0\\
        Omnet & 95.01 & 11.20 & 0.0\\
        Predator & 100.81 & 30.37 & 0.0\\
        CoFiNet & 99.34 & 16.97 & 0.0\\
        RoiTr & 16.79 & 3.10 & 43.1\\
        VRHCF & 110.62 & 16.35 & 0.0 \\
        GeoTrans & 18.15 & 1.69 & 81.7\\
        \midrule 
        Cross3DReg (Ours) & \textbf{6.68} & \textbf{1.01} & \textbf{87.1}\\
        \bottomrule
    \end{tabular}
    \label{tab:all_result} 
    \caption{The registration result on the Cross3DReg dataset.} 
    
\end{table}
\textbf{Quantitative Comparison}. In order to validate the effectiveness of the proposed method and assess the registration accuracy on the Cross3DReg dataset, we conduct a comparison experiment with the current mainstream point cloud registration methods, shown in Table $2$. The comparison methods ranges from a traditional iterative optimization method (ICP \cite{besl1992method}), feature learning-based registration methods (FCGF \cite{choy2019fully}, OMNet \cite{xu2021omnet}, VRHCF \cite{zhao2024vrhcf}), correspondence based methods ( Predator \cite{huang2021predator}), and coarse-to-fine registration frameworks (CoFiNet \cite{yu2021cofinet}, Roitr \cite{yu2023rotation}, GeoTrans \cite{qin2023geotransformer}).
The results show that our approach significantly outperforms the state-of-the-art methods in all evaluation metrics. In terms of registration accuracy, the lowest RRE = $6.68^\circ$ and RTE = $1.01$m are achieved, which reduces the rotation and translation error by $63.2\%$  and $40.2\%$, respectively. Our method also achieves the highest RR metrics, which is $5.4\%$ higher than the state-of-the-art GeoTrans.

The results also show that ICP is not able to handle significant point distribution differences in the case of cross-source point cloud registration. 
Additionally, although methods such as FCGF, OMNet, Predator, and CoFiNet perform well on the same-source point cloud reference, the RR is close to zero in the cross-source cases, showing that it is hard for the methods designed for same-source registration to handle point distribution variance and noise interference existing in cross-source point cloud registration. 

\begin{figure*}[ht]
\centering
\includegraphics[width=1\textwidth]{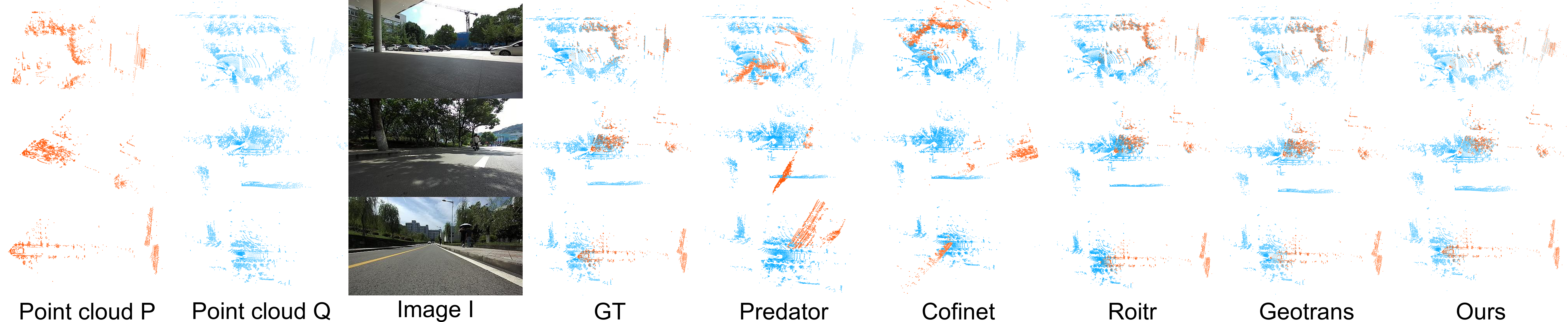} 
\caption{The visual registration results on Cross3DReg.}
\label{fig:registration}
\end{figure*}

\begin{figure*}
    \centering
    \includegraphics[width=1\linewidth]{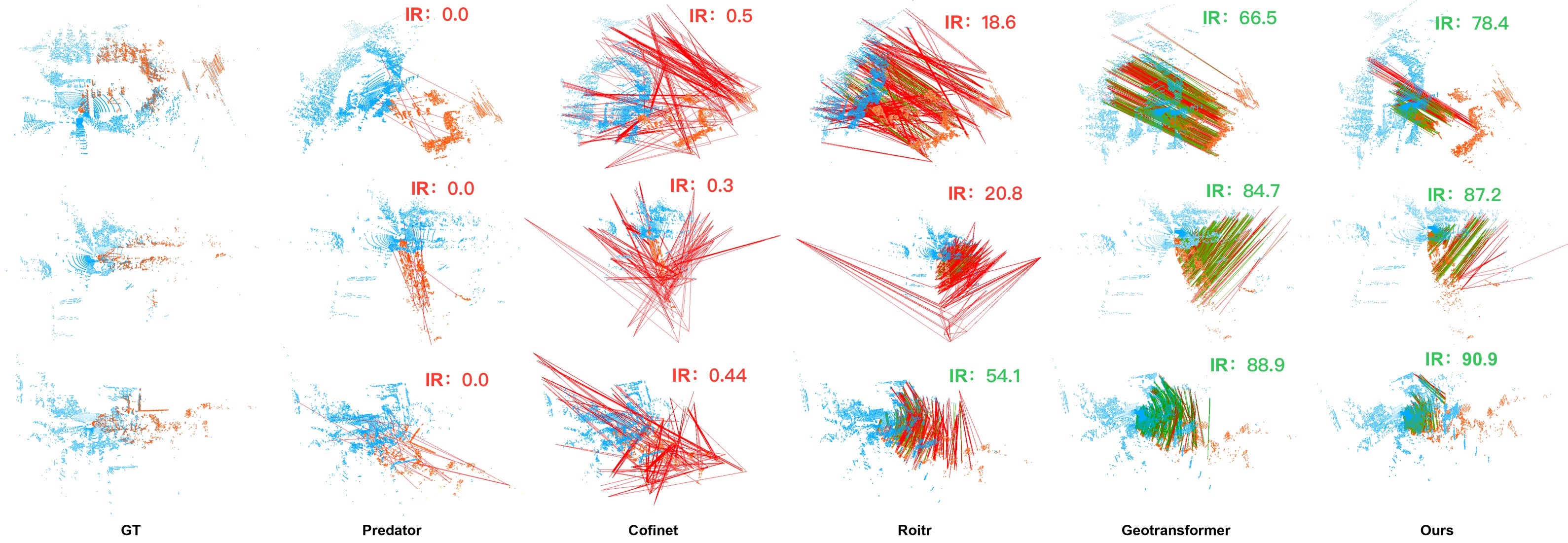}
    \caption{The Visualization of correspondences on Cross3DReg.}
    \label{fig:correspondencesl}
\end{figure*}
\begin{figure*}[!ht]
    \centering
    \includegraphics[width=1\linewidth]{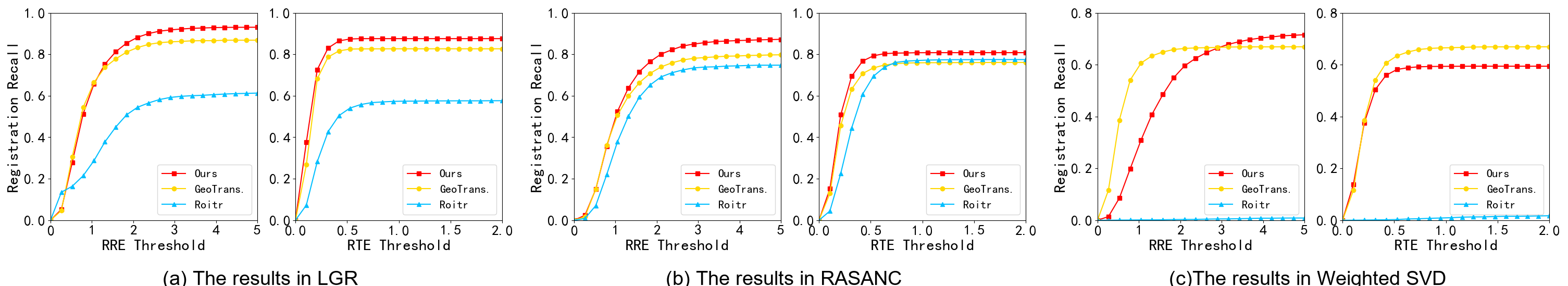}
    \caption{Registration recalls with different RRE and RTE thresholds in different estimators on Cross3DReg. }
    \label{fig:different_estimatorl}
\end{figure*}

To fully evaluate the correspondence of our cross-source point cloud registration method, we evaluate the registration accuracy with different estimators. As shown in Table \ref{tab:comparison}, when combined with the LGR \cite{qin2023geotransformer} pose estimator, our method exhibits optimal performance in all key evaluation metrics.
\begin{table}[!h]
\centering
\small
\begin{tabular}{c|l|llll}
\toprule
Estimator & Method & RRE $\downarrow$ & RTE $\downarrow$ & RR $\uparrow$ & IR $\uparrow$ \\
\midrule
\multirow{3}{*}{LGR} & Geotransr & 18.148 & 1.690 & 81.7 & 60.1 \\
 & RoiTr & 16.718 & 3.098 & 53.1 & 25.8 \\
 & Cross3DReg & \textbf{6.683} & \textbf{1.010} & \textbf{87.1} & \textbf{70.3} \\
\midrule
\multirow{5}{*}{RANSAC-50K} & RoiTr & 9.731 & 2.923 & 67.8 & 25.8 \\
 & Predator & 100.810 & 30.371 & 0.0 & 0.25 \\
 & CoFiNet & 99.342 & 16.973 & 0.0 & 0.13 \\
 & Geotrans & 13.05 & 1.436 & 72.2 & 60.1 \\
 & Cross3DReg & \textbf{8.150} & \textbf{1.135} & \textbf{78.6} &  \textbf{70.3} \\
\midrule
\multirow{3}{*}{Weighted SVD} & Geotrans & 16.491 & 2.193 & \textbf{62.9} & 60.1 \\
 & RoiTr & 54.210 & 16.541 & 0.2 & 25.8 \\
 & Cross3DReg & \textbf{8.684} & \textbf{1.153} & 57.8 &  \textbf{70.3} \\
\bottomrule
\end{tabular}
\caption{Registration results on the Cross3DReg dataset in different pose estimators. 
}
\label{tab:comparison}

\end{table}
When we employ the RANSAC estimator, our method still outperforms with RRE = $8.150^\circ$ and RTE = $1.135$ m, significantly better than GeoTrans, Predator, and CoFiNet, which also employ RANSAC. It is worth noting that both Predator and CoFiNet have an RR of $0.0\%$ in this setup, which indicates that the correspondences they generate are too low quality to support the RANSAC estimator to accomplish accurate alignment. 
Finally, we perform the tests in the weighted SVD \cite{besl1992method},
our method still achieves accurate  registration where RRE=$8.684^\circ$, RTE=$1.153$ m, and registration recall reaches $57.8\%$. In contrast, the registration recall of RoiTr is only $0.2\%$. Furthermore, it is observed that despite our method achieving approximately $10\%$ higher inlier ratio and lower registration error compared to GeoTrans, there is a gap in registration recall. The core reason for this is not due to the low quality of our generated correspondences, but rather the insufficient robustness of the pose estimator. Specifically, SVD, a least-squares solution, attempts to fit all given corresponding points. This makes it highly sensitive to the mismatched correspondences. 

\textbf{Qualitative Comparisons.}
Besides the quantitative comparison, in Figure \ref{fig:registration}, we also present a qualitative comparison of our method against current baseline approaches in the selected three challenging scenes, which has significant differences in density and pattern. 
As can be observed, methods such as Predator and CoFiNet fail to achieve successful alignment in the three challenging scenarios. In contrast to Roitr and GeoTransformer, our method attains the optimal registration performance, with results that are visually closest to the Ground Truth.

Figure \ref{fig:correspondencesl} visualizes the point correspondences generated by different registration approaches. The qualitative compar
ison clearly reveals that, compared to current State-of-the-art methods, the correspondences extracted by our method are more precise and highly concentrated on the overlapping regions of the point clouds. In contrast, methods like Predator, CoFiNet, and Roit produce a substantial number of incorrect correspondences. We attribute this primarily to two factors: a significant decrease in feature consistency within the overlapping regions when faced with considerable density variations across point clouds, and severe interference in feature matching caused by prominent noise in the scenes.

Furthermore, Figure \ref{fig:different_estimatorl} displays the registration recall of each method under different relative rotation and translation error thresholds, utilizing different pose estimators. The results demonstrate that, when robust pose estimators such as LGR and RANSAC are employed, the poses computed from the correspondences generated by our method exhibit the highest registration recall across all error thresholds.
\vspace{-0.136cm}
\subsection{Ablation Studies}
We conduct ablation experiments on the Cross3DReg dataset to evaluate the effectiveness of the module proposed in the method. As shown in Table \ref{tab:ablation}.
        
            
            
            
            
\begin{table}[!h]
    \centering
    \resizebox{\linewidth}{!}
    {
        \begin{tabular}{@{}l|cccc@{}}
            \toprule
            Method & RRE($^\circ$) & RTE(m) & RR($\%$) & IR($\%$) \\
            \midrule
            (a) Geo self-attention w/o OMP & 18.184 & 1.690 & 81.7 & 60.1 \\
            
            (b) OMP w/ Vanilla self-attention & 8.496 & 1.242 & 85.3 & 70.0 \\
            
            (c) OMP w/ Geo self-attention & 8.371 & 1.287 & 86.7 & 70.1 \\
            
            (d) OMP w/ VGAM(full) & \textbf{6.683} & \textbf{1.010} & \textbf{87.2} & \textbf{70.3} \\
            \bottomrule
        \end{tabular}
    }
    \caption{The registration result on the Cross3DReg dataset.}
    \label{tab:ablation}
\end{table}
Experiments are set up with four scenarios in comparison: (a) only the geometric self-attention module is used; (b) the OMP module is used with the vanilla attention module; (c) the OMP module is used with the geometric self-attention module; and (d) the complete Cross3DReg method. 
The comparison between (a) and (c) shows the OMP module can effectively mitigate the interference of redundant and noisy points, significantly improving the accuracy of cross-source point cloud alignment. The comparison of schemes (b), (c), and (d) further shows that our visual-geometric feature attention guidance module improves the consistency of feature space among cross-source point clouds, enhancing the alignment accuracy.

\section{Conclusion}
In this paper, we first propose a large-scale and real-world cross-source point cloud registration dataset, named Cross3DReg, to show different levels of noise, outliers, densities and structural patterns. An overlap-based cross-source point cloud registration method is then proposed to achieve accurate registration by predicting the overlap region with the help of unaligned images and ignoring redundant points and noise. 
Extensive experiments verify challenges of our proposed dataset and the effectiveness of the proposed method.

\bibliography{reference}

\end{document}